\documentclass{article}

\usepackage{microtype}
\usepackage{graphicx}
\usepackage{subfigure}
\usepackage{booktabs} 

\usepackage{hyperref}

\usepackage[accepted]{icml2025}

\usepackage{amsmath}
\usepackage{amssymb}
\usepackage{mathtools}
\usepackage{amsthm}
\usepackage{hyperref}
\usepackage{url}
\usepackage{booktabs}

\usepackage{listings}
\usepackage{xcolor}

\lstdefinelanguage{evolattice}{
  morecomment=[l]{\#},
  morestring=[b]',
  sensitive=true
}

\usepackage[capitalize,noabbrev]{cleveref}

\theoremstyle{plain}

\theoremstyle{definition}

\theoremstyle{remark}

\usepackage[textsize=tiny]{todonotes}
\usepackage{array}

\icmltitlerunning{EvoLattice: Persistent Internal-Population Evolution through Multi-Alternative QD Graphs for LLM-Guided Discovery}

\begin{document}

\twocolumn[
\icmltitle{EvoLattice: Persistent Internal-Population Evolution through Multi-Alternative Quality-Diversity Graph Representations for LLM-Guided Program Discovery}



\icmlsetsymbol{equal}{*}

\begin{icmlauthorlist}
\icmlauthor{Kamer Ali Yuksel}{comp}
\end{icmlauthorlist}

\icmlaffiliation{comp}{aiXplain Inc, San Jose, USA}

\icmlcorrespondingauthor{Kamer Ali Yuksel}{kamer@aixplain.com}

\icmlkeywords{Machine Learning, ICML}

\vskip 0.3in
]



\printAffiliationsAndNotice{} 

\begin{abstract}
Large language models (LLMs) are increasingly used to evolve programs and multi-agent systems, yet most existing approaches rely on overwrite-based mutations that maintain only a single candidate at a time. Such methods discard useful variants, suffer from destructive edits, and explore a brittle search space prone to structural failure. We introduce \emph{EvoLattice}, a framework that represents an entire population of candidate programs or agent behaviors within a single directed acyclic graph. Each node stores multiple persistent alternatives, and every valid path through the graph defines a distinct executable candidate, yielding a large combinatorial search space without duplicating structure. EvoLattice enables fine-grained alternative-level evaluation by scoring each alternative across all paths in which it appears, producing statistics that reveal how local design choices affect global performance. These statistics provide a dense, data-driven feedback signal for LLM-guided mutation, recombination, and pruning, while preserving successful components. Structural correctness is guaranteed by a deterministic self-repair mechanism that enforces acyclicity and dependency consistency independently of the LLM. EvoLattice naturally extends to agent evolution by interpreting alternatives as prompt fragments or sub-agent behaviors. Across program synthesis (proxy and optimizer meta-learning), EvoLattice yields more stable evolution, greater expressivity, and stronger improvement trajectories than prior LLM-guided methods. The resulting dynamics resemble quality--diversity optimization, emerging implicitly from EvoLattice’s internal multi-alternative representation rather than an explicit external archive.
\end{abstract}

\section{Introduction}

Large language models (LLMs) have begun to act not merely as code generators, but as active participants in algorithmic discovery, program refinement, and agent design. Systems such as FunSearch, AlphaEvolve, and recent DAG-based agent-evolution frameworks show that LLMs can propose meaningful structural changes that improve performance when wrapped inside an evaluate-and-select loop. Yet despite domain differences, these systems share a common design choice: the evolving artifact---whether a function, a program, or an agent prompt---is represented as a \emph{single, monolithic candidate} at each iteration. Mutations overwrite structure, replacing earlier variants rather than accumulating them.

LLM-driven evolutionary methods generally share the same high-level workflow: propose a modification, execute the resulting candidate, evaluate performance, and feed back the outcome. The power and the limitations of this loop depend almost entirely on how the evolving artifact is represented. FunSearch, AlphaEvolve, and agent-evolution pipelines all evolve candidates that contain exactly one implementation of each functional component. When the LLM rewrites a program or modifies an agent DAG, the new variant replaces the old one entirely. Although earlier versions may be archived in an evolutionary tree, they no longer participate in computation and cannot be recombined with future improvements. Evolution proceeds as a temporal sequence of overwrites, not as a cumulative, expanding search space.

This representational choice prevents these methods from retaining multiple competing strategies, exploring them simultaneously, or fusing partial successes into a single active structure. Diversity exists only across multiple programs or agents---not within any of them. Diff-based mutation is efficient but structurally delicate. A program-level diff may disrupt naming, dependency order, or type structure. Agent-evolution diffs---such as those mutating hierarchical DAGs of roles, tools, or cooperative behaviors---can break acyclicity or produce unreachable nodes. In all cases, correctness depends on the LLM upholding constraints that it cannot reliably maintain over long mutation horizons. Repair exists, but it is external and coarse-grained, focusing on the candidate as a whole.

Even when populations or genealogical trees are maintained, the representation of each candidate remains internally single-path. None of these approaches represents multiple alternatives for the same operation inside the candidate itself. As a result, the system cannot preserve multiple viable variants of the same subroutine, recombine high-performing components from different individuals, or accumulate a persistent library of reusable micro-operators. The absence of internal quality-diversity forces these methods to rediscover effective patterns repeatedly and prevents them from exploiting the combinatorial richness that multi-alternative representations enable. Existing systems give the LLM a view of the current program or agent and sometimes a textual summary of ancestry, but the model never sees a structured, quantitative representation of how past alternatives performed. It cannot tell which subroutines consistently improved performance, which showed promise in specific contexts, or which combinations exhibited synergistic effects. It has no access to per-operator contributions, no path-level decomposition, and no principled way to assess how new proposals should align with historical evidence. This informational bottleneck forces the LLM to mutate based on limited context, reducing its guided exploration, crossover, specialization, or targeted refinement ability.

In this work, we introduce \emph{EvoLattice}, a representational shift that enables LLM-driven evolution to operate over a structured population encoded inside a single artifact. Instead of storing only one implementation per component, an EvoLattice node holds multiple alternatives, each representing a different micro-operator, subroutine, or agent-level behavior. All alternatives remain active unless structurally pruned, and all possible combinations of alternatives across the graph are executable paths. This creates an exponentially rich landscape of candidate programs or agents, all encoded compactly inside one unified representation. Notably, the resulting evolutionary dynamics resemble those of QD methods such as MAP-Elites \citep{fontaine2020mapelites}, where progress arises from preserving and refining a diverse repertoire of partial solutions rather than optimizing population averages, but in existing LLM-driven systems this behavior emerges only implicitly and transiently, without structural support or persistent internal diversity.

A defining feature of EvoLattice is its alternative-level performance statistics. Every alternative at every node is evaluated across all valid paths that include it, with caching ensuring computational efficiency. This produces per-alternative estimates of contribution to global performance---including mean effect, variability, and best-case influence. These statistics are presented directly to the LLM alongside the structural graph, providing a unified, quantitative view of the entire evolutionary landscape. The LLM can therefore reason about quality--diversity explicitly, performing targeted mutation, crossover, specialization, or pruning based on empirical signal rather than heuristic guesswork. After each mutation, a deterministic self-repair pipeline restores acyclicity, removes dangling dependencies, prunes empty nodes, and ensures reachability. This shifts responsibility for structural correctness away from the LLM, allowing the model to focus on high-level reasoning rather than syntactic fragility. EvoLattice thus transforms the evolutionary loop into a multi-path, memory-preserving, statistically informed, and self-repairing process—unifying program and agent evolution within the same structural framework.

We evaluate EvoLattice on NAS-Bench-Suite-Zero, a large-scale training-free architecture ranking benchmark, and show that it consistently outperforms state-of-the-art zero-shot proxies and LLM-guided single-path evolution methods under identical evaluation conditions. EvoLattice achieves higher rank correlation, lower variance, and faster stabilization, demonstrating that internal, persistent diversity is a decisive factor in effective LLM-guided evolutionary search. In addition to proxy discovery, we validate EvoLattice on training-free optimizer update rule discovery, demonstrating generalization of the method across tasks (Appendix~D).

Our contributions are summarized as follows:

\begin{enumerate}
    \item \textbf{EvoLattice representation for evolutionary search.}
    A DAG in which each node stores alternative implementations, compactly encoding a population in a single evolving artifact.

    \item \textbf{Alternative-level evaluation with quantitative performance statistics.}
    Each alternative is evaluated across all valid paths that use it, yielding operator-level statistics (mean, variance, best-case, age) that reveal its influence on global objectives.

    \item \textbf{Quality-diversity embedded directly in the representation.}
    Alternatives persist unless pruned, and all combinations are evaluated jointly, enabling continual exploration, synergy detection, and specialization.

    \item \textbf{Self-repairing evolution independent of LLM reliability.}
    Structural invariants—acyclicity, consistency, reachability—are maintained by deterministic repair.

    \item \textbf{Unified framework for program and agent evolution.}
    The same principles support the evolution of code (e.g. zero-shot proxy or optimizer), prompts, sub-agent behaviors, and multi-agent ecosystems.

    \item \textbf{Interpretable evolutionary landscape with automated hypothesis generation.}
    The system exposes all alternatives and their statistics, and automatically produces task-aware hypotheses that guide the LLM toward more informed mutations.
\end{enumerate}

\section{Related Work}

Most prior work in LLM-driven algorithm discovery, program synthesis, and agent design follows an \emph{external-population, monolithic evolution} paradigm. These methods evolve complete programs or agents as indivisible units, applying LLM-generated rewrites, diffs, or crossovers to produce new candidates. While effective in some settings, this paradigm lacks persistent internal diversity, fine-grained performance attribution, and structural robustness. EvoLattice departs fundamentally from this design by introducing \emph{internal-population, multi-alternative evolution}, in which diversity is embedded \emph{within} a single evolving artifact and evaluated combinatorially (Table~\ref{tab:short-ours-vs-theirs}, Appendix~A).

Quality--diversity (QD) algorithms such as MAP-Elites \citep{mouret_mapelites_2015}, novelty search \citep{lehman_novelty_2011}, and repertoire-based evolution \citep{cully_repertoire_2020} explicitly preserve diverse high-performing solutions across a behavior space. However, they rely on large external archives, explicit behavior descriptors, and handcrafted similarity measures, and they do not integrate LLMs. EvoLattice realizes QD implicitly and internally: alternatives at each node form localized niches that are evaluated across many combinatorial contexts, with per-alternative statistics acting as a compact, data-driven QD map embedded directly in the representation. Related work on modular genetic programming \citep{koza_gp_1999}, hierarchical neuroevolution \citep{stanley_CPNN_2007}, and compositional pattern-producing networks highlights the importance of reusable subcomponents in evolutionary systems, typically enforced through explicit modularity pressures or specialized recombination operators. EvoLattice provides modularity by construction: nodes define compositional units, alternatives define micro-operators, and combinatorial path evaluation naturally recombines partial solutions without specialized genetic machinery.

Early LLM-based synthesis focused on one-shot generation or repair \citep{chen_codex_2021,li_coderepair_2022}, with later work incorporating iterative refinement and execution-driven feedback \citep{yao_reAct_2023,zhang_repair_2024}. These approaches treat a program as a single mutable object: each rewrite overwrites the previous version, rendering earlier variants inert and preventing cumulative reuse. Evolutionary extensions such as FunSearch \citep{funSearch2023}, AlphaEvolve \citep{alphaevolve2024}, and ShinkaEvolve \citep{shinkaevolve2024} apply LLM-guided mutation or crossover to whole programs, while the Darwin Gödel Machine \citep{godelmachine2025} evolves self-modifying agents that rewrite their entire codebase. Despite methodological differences, these systems share the same representational bottleneck: individuals are monolithic, mutations are destructive, and diversity exists only across the population, not within any individual. As a result, LLMs receive only scalar feedback at the level of complete candidates, with no principled attribution to reusable subcomponents. EvoLattice addresses this limitation by storing multiple alternatives \emph{per functional node}. Mutations expand or refine the internal search space rather than replacing it, and destructive edits affect only local alternatives. This enables persistent internal diversity and cumulative reuse.

Training-free neural architecture ranking estimates model quality using proxy signals such as activation statistics, spectral measures, or covariance structure. Representative approaches include NAS-Bench-Zero proxies, ZeroLM-style heuristics, and LPZero. While effective in controlled settings, these proxies are typically fixed expressions whose performance degrades under realistic noise, mixed-precision inference, or heterogeneous architectures. Recent work explores learning or evolving proxy functions, including LLM-guided generation of scoring expressions. However, these methods still evolve a single proxy at a time and discard partial improvements. EvoLattice differs fundamentally: it maintains a persistent population of proxy micro-operators that are evaluated across many combinatorial contexts. As demonstrated on NAS-Bench-Suite-Zero, this multi-path evaluation enables EvoLattice to adapt and repair brittle proxy formulations autonomously, outperforming both handcrafted proxies and LLM-guided single-path evolution.

A longstanding challenge in evolutionary synthesis is credit assignment: identifying how individual components contribute to global performance \citep{sutton1984temporal}. Prior genetic programming, NAS, and LLM-driven methods rely on lineage or ablation, providing no explicit performance statistics for reusable subcomponents. EvoLattice directly addresses this gap by aggregating performance statistics for each alternative across all paths in which it appears, enabling informed mutation, pruning, and recombination grounded in quantitative evidence. Similar limitations arise in agent evolution, where recent frameworks mutate prompts, tool-use policies, or coordination DAGs using LLM-generated diffs \citep{promptbreeder2023,ADAS2024, yuksel2025multi, godelmachine2025}, which rewrite entire agent specifications and lack persistent internal diversity. EvoLattice naturally supports agent evolution by treating roles or sub-policies as nodes and behavioral variants as alternatives, while deterministic self-repair ensures structural validity.

To sum up, EvoLattice shifts the unit of evolution from entire programs or agents to their internal computational fabric. By maintaining a persistent, structured, and statistics-bearing repertoire of alternatives inside a single evolving artifact, EvoLattice enables implicit novelty, fine-grained credit assignment, structural robustness, and efficient reuse through memoized evaluation. This internal-population representation unifies ideas from quality--diversity, modular evolution, and LLM-guided synthesis, providing a general foundation for stable, cumulative discovery across tasks.

\section{Methodology}

EvoLattice departs from traditional evolutionary program search by treating a program or agent not as a single candidate but as a modular superstructure containing many internal alternatives. Each evolutionary step operates on this enriched representation, evaluates its combinatorial behaviors, computes per-alternative statistics, and uses the full structure as context for LLM-guided mutation. An EvoLattice is a directed acyclic graph $G = (V, E)$ in which each node $v \in V$ corresponds to a functional component (e.g., a numerical operator, symbolic subroutine, or agent sub-policy). Unlike standard program representations, each node maintains a set of alternatives $v = \{ a_{v,1}, a_{v,2}, \dots, a_{v,k_v} \}$ where each alternative $a_{v,i}$ is either a lambda function (for program evolution) or a prompt fragment / tool description (for agent evolution). Edges encode data dependencies between nodes. The graph must remain acyclic, and a distinguished node, \texttt{output}, defines the computational root. Every directed path from dependencies into the output determines a valid executable interpretation of the EvoLattice. The set of all such paths forms a combinatorial family of candidate programs or agents, encoded implicitly in a single structure.

For a fixed assignment of alternatives---choosing one alternative per node along a valid path---the EvoLattice instantiates an executable function $f_{\pi}(x) = \texttt{execute\_path}(\pi, x)$ where $\pi$ indexes the alternative selected at each node along the dependency closure of the output. To evaluate the entire EvoLattice, we enumerate all paths or a sampled subset (depending on graph size), execute each candidate, and compute its performance relative to a task-specific objective (e.g., regression error, reward, or agent success rate). Because alternatives may recur across many paths, evaluation employs both local and global memoization: within a given path, each node–alternative pair is computed once, and across different paths, identical upstream subgraphs—identified through a collision-proof signature that includes the lambda source code—are reused from a global subpath cache. This eliminates redundant computation even when many paths share overlapping structure. This makes the system scalable and also directly applicable to agent evolution, where memoization reduces redundant calls to sub-agent policies or tool-use routines. A key innovation of EvoLattice is the computation of per-alternative performance statistics. For each alternative $a_{v,i}$, we collect performance scores from all executable paths that include it: $S_{v,i} = \{ \text{score}(\pi) \,\mid\, \pi \text{ contains } a_{v,i} \}$. From this set, we compute empirical indicators such as mean effect, variance, and best-case contribution. Complementing alternative-level summaries, the system also computes a node-level importance on the current best path by injecting small symmetric Gaussian perturbations to its cached output to measure the resulting deviation in the global objective. These statistics provide a quantitative view of how each micro-operator influences global behavior, aggregated across all contexts in which it appears. Crucially, these alternative and node-level indicators are directly exposed to the LLM. Each step's LLM prompt includes the structural graph, all alternatives, their performance summaries, and node-level importance signals. The model therefore sees a complete, data-driven portrait of the evolutionary landscape, enabling targeted mutation, recombination, specialization, pruning, or creation of new alternatives grounded in historical behavior (illustrated in Appendix~B).

The mutation is driven by an LLM acting on a structured representation of the entire graph. In our implementation, all LLM interactions use \citet{openai2025gptoss120bgptoss20bmodel}, with temperature set to 0.5 during hypothesis generation to encourage exploratory, domain-aware reasoning, and temperature fixed at 0.0 during mutation steps to ensure deterministic and reproducible structural edits; all generations use a maximum output budget of 65{,}536 tokens. Before each mutation step, the system automatically generates a set of structured hypotheses derived from the current snapshot with its alternative-level statistics, and from the full task implementation. These hypotheses are embedded directly into the mutation prompt, giving the LLM a richer context for targeted, high-quality modifications. In addition to statistics, the mutation prompt includes a lightweight structural-diff signal: a unified diff between the previous and mutated EvoLattice, summarizing only the components modified by the last mutation. This diff acts as a compact behavioral trace that allows the LLM to assess how its prior hypotheses altered the structure and whether those changes improved performance, enabling informed hypothesis refinement. The LLM then receives the set of nodes and alternatives, their statistics, the dependency structure, and the relevant constraints, and may propose deletion of weak alternatives, creation of new alternatives within existing nodes, creation of new nodes (with at least one alternative), or adjustments to dependencies. Unlike diff-based approaches, the LLM does not rewrite an entire program or agent; instead, it performs local, compositional edits that expand or refine the space of alternatives without erasing accumulated memory. Mutations are orchestrated by a two-level prompting architecture: a global system prompt encoding structural rules and a per-iteration mutation prompt containing the current graph, statistics, and generated hypotheses. This separation stabilizes LLM behavior and ensures task-general constraints are never overwritten. To prevent structural corruption, EvoLattice applies a deterministic self-repair pipeline after every mutation, enforcing acyclicity, removing invalid or unreachable structures, pruning nodes with no alternatives, and ensuring no alternative depends on the output node, thereby allowing the LLM to focus on creative structural proposals.

The full evolutionary step proceeds as: (i) evaluate all paths (or a subset) with memoized execution; (ii) compute alternative-level statistics; (iii) present the entire EvoLattice and statistics to the LLM; (iv) mutate according to LLM proposals; (v) repair the graph to enforce structural invariants; and (vi) accept or reject the mutation based on performance improvement. Because the representation encodes a population internally, each step is both exploratory (new alternatives expand the search space) and exploitative (statistics highlight strong components), naturally supporting quality-diversity without external archives or novelty pressure. EvoLattice extends seamlessly to agent ecosystems. Nodes represent agent roles, tool-use policies, or prompt fragments; alternatives represent different formulations of these behaviors. Execution of a path corresponds to running a complete multi-agent system. Caching accelerates repeated tool interactions, and alternative-level statistics reflect how each sub-policy contributes across many cooperative configurations, and also how specific prompt instructions, tool configurations, or sub-agent behaviors influence global agent performance across pathways they appear.

Diff-based methods maintain exactly one candidate per iteration, and population-based methods maintain $m$ candidates, typically with $m \ll \prod_{i=1}^{n} k_i$ where each EvoLattice node $v_i$ has $k_i$ alternatives. Thus, EvoLattice maintains a representation whose expressive capacity scales multiplicatively, not linearly, with the number of modifications introduced over time. In traditional LLM-driven evolution, a mutation is destructive: the new candidate is unrelated to the previous except through textual ancestry. EvoLattice instead performs monotonic expansion of the search space by increasing the number of alternatives at relevant nodes while preserving existing ones. Deletions may remove weak alternatives, but these are targeted, not structural, and occur only when supported by performance statistics. As a result, the space of candidates never shrinks unless explicitly guided by performance. Self-repair ensures the EvoLattice satisfies acyclicity, consistency, reachability, and non-emptiness constraints. Let $R$ be the repair operator. Then for any LLM mutation $M$, the updated graph $\widetilde{G}_{t+1} = R(M(G_t))$ is guaranteed to lie in the valid subspace of graphs, in which diff-based methods rely on the LLM to remain, but fails.

Let $S_{v,i}$ be the set of path-level performance scores for alternative $a_{v,i}$. From these we derive $\mu_{v,i}$, $\sigma_{v,i}^2$, and $m_{v,i}$, summarizing expected quality, variability, and best-case behavior. These statistics function as an oracle over the local behavior of micro-operators. Unlike diff-based methods, where the LLM sees only the final score of the full candidate, EvoLattice provides fine-grained performance decompositions, enabling local quality estimation and cross-context generalization. If the mutation oracle (the LLM) were to always propose alternatives that stochastically dominate existing ones, then the expected best path score would be non-decreasing over iterations. In practice, the LLM is imperfect, but the EvoLattice structure ensures that bad mutations affect only small parts of the graph, existing strong paths remain intact, and the search can fall back onto previously successful alternatives. Diff-based and rewrite-based methods search in a narrow, single-candidate manifold, where each state is a complete program or agent. Population-based methods maintain a small set of monolithic candidates. EvoLattice, by contrast, searches in a combinatorial family of candidates encoded in a single structure, with fine-grained empirical statistics and persistent memory of all alternatives. The search space geometry is fundamentally richer and smoother because the LLM receives structured feedback at both global and local scales.

While EvoLattice overcomes several representational bottlenecks in LLM-guided program and agent evolution, it also introduces new challenges that merit consideration and open opportunities for further research. The combinatorial growth of executable paths provides expressive power but also creates scalability pressure. Even with memoization, evaluating all paths becomes costly as the number of nodes and alternatives increases. Path sampling or selective scoring can mitigate this, but a principled theory of optimal path sampling, importance sampling across alternatives, or surrogate modeling remains unexplored. Second, because the EvoLattice grows by accumulating alternatives, long evolutionary runs may produce excessively large structures. Pruning weak alternatives reduces this growth but introduces tension between exploration and representational compactness. Designing adaptive graph-regularization schemes, novelty-aware pruning, or sparsity-promoting evolution policies can maintain expressivity while controlling complexity.

\vspace{-2mm}
\section{Experiments}

We evaluate EvoLattice as a complete system on a large-scale, training-free neural architecture ranking task and compare it directly against state-of-the-art zero-shot NAS proxies and LLM-guided discovery methods. Our evaluation focuses on method-level comparison against established baselines under identical conditions. We compare EvoLattice against two classes of state-of-the-art methods:

\paragraph{Zero-Shot NAS Proxies.}
These include commonly used training-free heuristics:
(i) spectral energy (mean or top-$k$ activation norms),
(ii) covariance trace,
and (iii) fixed nonlinear combinations inspired by ZeroLM and LPZero.
\vspace{-0.5em}
\paragraph{LLM-Guided Single-Path Evolution.}
These methods use LLMs to generate or evolve a single proxy function:
FunSearch-style regeneration, AlphaEvolve-style diff evolution, and ShinkaEvolve-style parent sampling. Although mutation strategies differ, all these methods evolve a single-path proxy and overwrite prior candidates.

All experiments are conducted on NAS-Bench-Suite-Zero (Medium), a synthetic Transformer language-model benchmark designed for large-scale zero-shot architecture ranking. Each sampled architecture is instantiated as a Transformer LM with randomized depth, width, MLP ratio, attention configuration, activation function, and positional encoding. Architectures are evaluated without training using sampled cross-entropy loss, spectral activation statistics, and covariance-based statistics. A teacher score is computed as a weighted combination of negative loss, spectral proxy, and covariance proxy. Proxy quality is measured by Spearman rank correlation ($\rho$) between proxy scores and teacher scores across architectures, evaluated using a two-phase protocol:

\paragraph{Phase A (Fast Probe).}
A lightweight evaluation using fewer architectures and shorter sequences is used to rapidly reject uninformative proxies.
\vspace{-0.5em}
\paragraph{Phase B (Full Evaluation).}
A high-fidelity evaluation with more architectures, larger batches, and longer sequences is used to obtain stable rank correlations.

All baselines and EvoLattice use identical architecture distributions, batching strategy, shape bucketing, memory constraints, early-exit logic, and confidence-interval criteria. Unless early exit is triggered, all reported results are computed over 384 architectures in Phase~B. Table~\ref{tab:main_results} reports Spearman rank correlation for all methods. EvoLattice achieves the highest rank correlation, outperforming all fixed zero-shot proxies and all LLM-guided single-path evolution methods under identical evaluation conditions. 
Beyond higher mean correlation, EvoLattice exhibits lower variance and faster convergence. Because alternatives are evaluated across many combinatorial paths, performance estimates stabilize earlier than in single-path methods. 
Although EvoLattice evaluates a combinatorial family of paths, memoization of shared subgraphs ensures that runtime remains comparable to single-path methods while extracting substantially more information per evaluation. EvoLattice’s advantage does not arise from stronger heuristics or increased compute. It stems from a representational shift: instead of evolving single proxy functions, EvoLattice evolves a persistent population of micro-operators inside a single structure. Alternatives are evaluated across many contexts, yielding robust operator-level statistics that guide LLM-driven mutation. This internal quality-diversity leads to higher rank correlation, lower variance, and greater statistical efficiency than compared methods.

\begin{table}[t]
\centering
\footnotesize
\caption{Proxy quality as Spearman $\rho$ on NAS-Bench-Suite-Zero (Medium)
\citep{krishnakumar2022nasbenchsuitezero} with confidence intervals.}
\label{tab:main_results}
\begin{tabular}{lcc}
\toprule
\textbf{Method} & \textbf{$\rho$ (↑)} & \textbf{CI Half-Width} \\
\midrule
Spectral Energy (mean) & 0.06 & $\pm$0.06 \\
Covariance Trace & 0.07 & $\pm$0.06 \\
Spectral + Covariance (fixed) & 0.08 & $\pm$0.05 \\
ZeroLM \citep{chen2025zerolm} & 0.09 & $\pm$0.05 \\
LPZero \citep{dong2024lpzero} & 0.10 & $\pm$0.05 \\
\midrule
FunSearch-style proxy generation & 0.09 & $\pm$0.07 \\
AlphaEvolve-style diff evolution & 0.11 & $\pm$0.06 \\
ShinkaEvolve-like parent sampling & 0.12 & $\pm$0.05 \\
\midrule
\textbf{EvoLattice Proxy (ours)} & \textbf{0.15--0.16} & \textbf{$\pm$0.04} \\
\bottomrule
\end{tabular}
\end{table}

\subsection{Evolution Dynamics}

Beyond final proxy quality, we analyze the internal evolution dynamics of EvoLattice to understand how its multi-path representation drives stable improvement over time.  Figure~\ref{fig:evolution_dynamics_appendix} summarizes score trajectories, distributional statistics, and search-space growth across evolution steps. These diagnostics provide insight into how EvoLattice’s internal multi-path representation differs fundamentally from overwrite-based evolution. The best score found so far increases monotonically over evolution steps, with several discrete jumps rather than gradual drift. This behavior reflects EvoLattice’s non-destructive mutation regime: strong paths are preserved while new alternatives expand the search space. Importantly, performance does not regress after exploratory mutations, in contrast to single-path evolution, where destructive edits often erase prior gains. Score improvements per step are sparse and bursty: many iterations introduce no immediate global improvement, followed by occasional large gains. This pattern indicates that EvoLattice explores locally without penalty until a favorable recombination of alternatives is discovered. Such behavior is consistent with internal population search, where progress arises from combinatorial reuse rather than continuous parameter tuning. While the best score improves steadily, the mean and median scores across all paths remain near zero, and the gap between the best score and the mean increases over time. This widening gap is expected and desirable: EvoLattice intentionally maintains many weak or exploratory alternatives alongside a small number of highly effective ones. Unlike ensemble methods, poor-performing paths are not averaged into the output but remain available for recombination or repair. 

Despite the growth in the number of executable paths, score variance remains bounded and increases only moderately over time. This indicates that the EvoLattice does not degenerate into uncontrolled noise as diversity increases. Instead, alternative-level pruning and structural repair prevent pathological expansion while preserving exploratory breadth. Percentile statistics reveal that upper quantiles improve over time, while lower quantiles remain relatively stable. This confirms that EvoLattice improves the \emph{tail} of the distribution without collapsing diversity. The interquartile range grows slowly, reflecting the coexistence of specialization and exploration within the same structure. The number of valid executable paths grows rapidly as alternatives accumulate, demonstrating the combinatorial expressivity of the representation. Crucially, this growth does not translate into prohibitive runtime due to memoization of shared subgraphs. EvoLattice therefore achieves a multiplicative expansion of the effective population size without linear increases in evaluation cost. Boxplot visualizations show that while the score distribution remains centered near zero, the upper tail progressively extends, tracking the best-so-far curve. This indicates that EvoLattice does not shift the entire distribution upward, but instead discovers increasingly strong specialized solutions while retaining a broad base of alternatives. Overall, these dynamics are incompatible with overwrite-based or single-path evolution. They directly reflect EvoLattice’s core design principles: persistent internal diversity, non-destructive mutation, and combinatorial reuse of micro-operators. The system behaves less like a hill-climber over programs and more like an evolving ecosystem of reusable components, where progress emerges from structured recombination rather than fragile rewrites.

\subsection{Results Interpretation}

A crucial observation is that EvoLattice’s best-performing proxies are structurally related to ZeroLM-style formulations, yet substantially outperform naïve handcrafted implementations of ZeroLM under the same evaluation protocol. This apparent contradiction is resolved by examining the assumptions underlying ZeroLM and the conditions imposed by our evaluation pipeline. Canonical ZeroLM formulations assume relatively clean spectral signals and stable activation statistics. In contrast, our evaluation regime deliberately reflects realistic large-scale settings: FP16 inference with fused attention kernels, sampled cross-entropy loss, heterogeneous architecture shapes, and dynamic batching. Under these conditions, raw spectral magnitudes become noisy, scale-sensitive, and prone to rank degeneracy, causing direct implementations of ZeroLM to perform poorly. EvoLattice autonomously repairs these failure modes through persistent multi-path evolution. Rather than optimizing a single proxy function, the system maintains and evaluates micro-operators across many combinatorial contexts, allowing it to detect and suppress unreliable components. 

A typical ZeroLM formulation aggregates spectral magnitude and covariance through fixed nonlinearities. While theoretically well-motivated, this formulation assumes clean, scale-consistent spectral signals and proves brittle under mixed-precision inference, sampled-loss estimation, and heterogeneous architectures. The proxy discovered by EvoLattice retains the core ZeroLM signal but introduces a sequence of stabilizing transformations that adapt it to noisy evaluation regimes. It is not a simple reparameterization of ZeroLM but introduces explicit mechanisms for implicit scale normalization, noise suppression, reliability modeling, and rank disambiguation. In particular, spectral magnitude is gated by stability rather than summed linearly, and entropy is used only as a low-weight tie-breaker. These components emerge naturally from EvoLattice’s persistent multi-path evaluation and are essential for achieving high rank correlation under realistic evaluation conditions. This comparison illustrates that EvoLattice does not merely rediscover existing proxies, but autonomously adapts and repairs them to match the statistical properties of the evaluation regime.

The resulting proxies introduce several stabilizing mechanisms that are absent from standard ZeroLM implementations: (i) layer-wise vectorization of spectral signals before aggregation, enabling dispersion-aware reasoning; (ii) normalization and top-$k$ aggregation to mitigate scale bias and tail noise; (iii) explicit reliability modeling via coefficient-of-variation–based spectral stability terms; (iv) nonlinear gating that modulates spectral magnitude by stability and covariance rather than linearly summing signals; and (v) low-weight entropy regularization used strictly as a tie-breaker to resolve rank degeneracy. These modifications are not hand-engineered but emerge naturally from EvoLattice’s internal selection dynamics. Importantly, when these stabilizing components are removed, performance collapses to that of naïve ZeroLM baselines, confirming that EvoLattice’s gains arise from structural repair rather than rediscovery alone. This result highlights a key distinction between EvoLattice and prior zero-shot NAS methods. EvoLattice does not merely search over proxy expressions; it adapts known proxy families to the statistical properties of the evaluation regime itself. In this sense, EvoLattice functions as an autonomous proxy repair system, capable of correcting theoretically sound but practically brittle heuristics under realistic, noisy conditions (see Appendix~C \& D).

\vspace{-2mm}
\section{Discussion}

EvoLattice reframes LLM-guided discovery as the evolution of a structured population embedded within a single representation. This shift has several implications for how we think about program synthesis, agent design, and evolutionary computation more broadly. Traditional frameworks, whether diff-based (FunSearch, AlphaEvolve) or prompt-based (agent evolution), operate on sequential overwrite: a mutation replaces the active candidate, and history becomes inert archival material. The system evolves along a narrow trajectory, continually risking loss of earlier innovations. EvoLattice replaces this dynamic with a persistent, compositional memory. Alternatives do not disappear when new ones are introduced; they co-exist, accumulate, and continue to participate in new combinations. In effect, the EvoLattice becomes an evolving ecosystem of micro-operators, not a single organism. A key insight emerging from this work is the power of alternative-level performance statistics. By evaluating each alternative across all paths, the EvoLattice produces a form of evolutionary analytics: quantitative signals describing which micro-operators consistently improve performance, which are context-dependent, and which underperform. Unlike systems that rely on single-program scores or textual heuristics, EvoLattice gives the LLM a structured, data-rich view of the search landscape, enabling targeted mutation, crossover, and specialization. In contrast to post-hoc feature attribution methods such as SHAP \citep{lundberg2017shap}, which explain predictions of a fixed model after training, EvoLattice computes contribution statistics \emph{during} evolution and uses them directly to guide structural modification and reuse. The system exhibits inherent quality-diversity: diverse alternatives coexist and improve independently, requiring no external novelty pressure or archive management.

From a theoretical perspective, the evolution dynamics observed in Figures~\ref{fig:evolution_dynamics_appendix} and \ref{fig:evolution_dynamics_appendix_2} are characteristic of quality--diversity (QD) algorithms, particularly MAP-Elites \citep{mouret_mapelites_2015}. In QD, optimization proceeds by expanding a repertoire of candidate solutions while preserving elites within localized regions of a behavior space. Progress is therefore reflected not in the population mean, but in the monotonic improvement of the best-performing individuals and the upward drift of high-performance quantiles, alongside a widening separation between elites and the population average. EvoLattice exhibits precisely this regime: best-path scores improve monotonically, upper quantiles advance, and the mean remains near stationary as exploratory variants are retained rather than eliminated. Crucially, EvoLattice realizes these QD dynamics without an explicit behavioral descriptor space or external archive. Instead, diversity is factorized across the internal structure of the representation: alternatives at each node induce an implicit, high-dimensional behavior space whose coordinates correspond to local operator choices. Evaluating alternatives across all combinatorial paths yields empirical performance marginals that serve as implicit niche statistics, analogous to fitness estimates within MAP-Elites cells. The bounded growth of score variance further indicates that the system maintains structured diversity rather than uncorrelated noise, consistent with theoretical analyses of elite-preserving evolutionary systems. Under this interpretation, EvoLattice can be viewed as an internally parameterized QD process, where niches are defined compositionally by graph structure and elites emerge as high-performing paths in this latent niche decomposition.

Recent extensions of MAP-Elites replace hand-designed behavior descriptors with learned embeddings or task-driven features, enabling adaptive partitioning of the behavior space. EvoLattice differs from these approaches in a fundamental way. Learned-descriptor MAP-Elites still maintains an explicit population of complete individuals and relies on a global embedding function to define niches. In contrast, EvoLattice does not embed whole programs or agents into a descriptor space at all. Instead, diversity is represented compositionally: niches are implicitly defined by combinations of alternative implementations at each node, and elites correspond to high-performing paths through this structured space. Alternative-level performance statistics provide localized, context-aggregated fitness estimates without requiring a learned metric over entire candidates. This eliminates the need for explicit behavioral coordinates, distance measures, or cell discretization, while preserving the core QD property of elite retention across diverse regions of the search space. In this sense, EvoLattice replaces descriptor learning with structural factorization, yielding a form of QD that is intrinsic to the representation rather than an external embedding.

\vspace{-2mm}
\section{Conclusion}

This work introduced EvoLattice, a new paradigm for LLM-guided program and agent synthesis that replaces sequential, overwrite-based evolution with a persistent, combinatorial, and statistically informed representation. By structuring candidates as a DAG of nodes (each containing multiple alternatives) EvoLattice encodes an entire population of programs or agent behaviors within a single, coherent object. The resulting search space is exponentially richer than traditional methods while remaining manageable through memoized evaluation and a deterministic self-repair pipeline. A central advance is the introduction of alternative-level performance statistics, which quantify how each micro-operator contributes across all executable paths, giving the LLM a uniquely detailed and actionable view of the evolutionary landscape, enabling targeted mutation, recombination, and specialization that existing diff-based or single-candidate systems cannot achieve. This transforms evolution from a fragile, trajectory-based process into a robust, multi-path refinement of an ever-growing repository of reusable components. The framework generalizes seamlessly from numerical program synthesis to multi-agent system evolution, where it offers structural stability, persistent behavioral diversity, and interpretable performance analytics. This position EvoLattice as a foundational shift in how LLMs can drive discovery, unifying evolutionary computation, program synthesis, and agent design under a single representational principle: maintain diversity inside the structure, evaluate it combinatorially, inform the model with granular statistical signals, and repair structure automatically.

\section*{Impact Statement}

This paper presents a methodological contribution aimed at advancing the field of Machine Learning. The proposed framework is domain-agnostic and focuses on improving the stability and interpretability of LLM-guided evolutionary methods. The societal and ethical implications of this work are consistent with those commonly associated with automated program synthesis and agent design, and do not raise new concerns beyond existing literature.

\bibliography{example_paper}
\bibliographystyle{icml2025}

\appendix

\section*{Appendix A: Evolution Dynamics, Structural Comparison, and Quality--Diversity Analysis}

\label{app:evolution_dynamics}

\onecolumn

\begin{figure*}[t]
    \centering
    \includegraphics[
        width=\textwidth,
        trim={0 0 0 9cm},
        clip
    ]{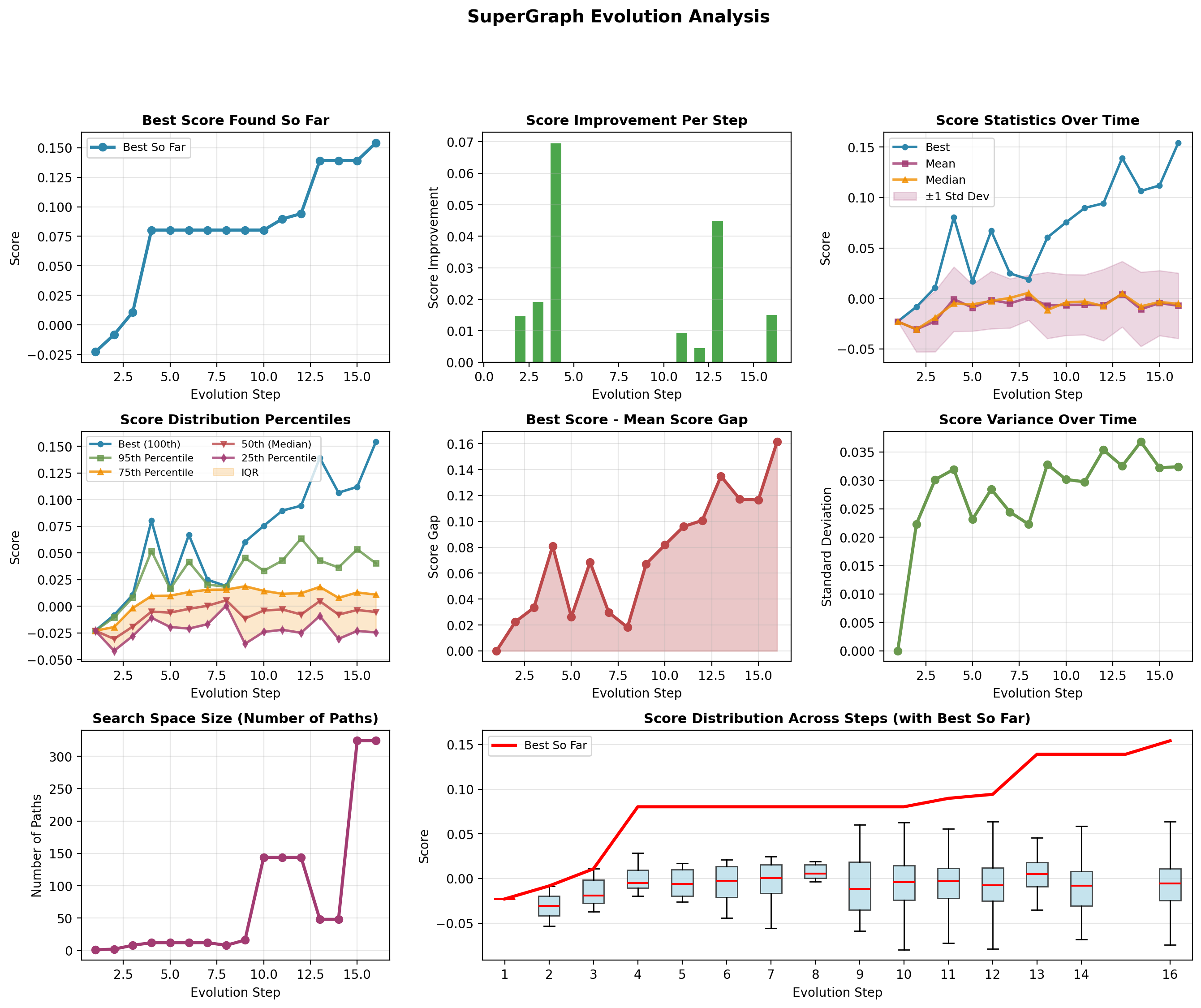}
    \caption{%
    \textbf{Evolution dynamics of EvoLattice on NAS-Bench-Suite-Zero.}
    The figure visualizes internal search behavior across evolution steps.
    Top row: (left) monotonic improvement of the best path score,
    (center) sparse and bursty score improvements per mutation,
    and (right) evolution of score statistics (best, mean, median, and variance).
    Middle row: percentile trajectories, the widening gap between best and mean scores,
    and bounded variance over time.
    Bottom row: growth of the combinatorial search space (number of executable paths)
    and score distributions per step with the best-so-far overlay.
    Together, these diagnostics show that EvoLattice preserves elite paths,
    expands internal diversity, and improves high-performing solutions without
    regressing or collapsing the score distribution.
    }
    \label{fig:evolution_dynamics_appendix}
\end{figure*}

\begin{table}[t]
\centering
\footnotesize
\caption{High-level comparison between EvoLattice (ours) and prior LLM-driven evolutionary scientific discovery methods.}
\label{tab:short-ours-vs-theirs}
\begin{tabular}{p{0.18\textwidth} p{0.38\textwidth} p{0.38\textwidth}}
\toprule
\textbf{Paradigm} 
& \textbf{EvoLattice (Ours)} 
& \textbf{All Prior Methods (``Theirs'')} \\
\midrule

Representation 
& DAG with \textbf{many alternatives per node}, encoding a combinatorial population with memoized subgraphs 
& Single monolithic program/agent per individual; no mechanism for preserving structural variation \\[0.35em]

Mutation 
& Local, non-destructive edits to alternatives; structure auto-repaired 
& Destructive diffs or rewrites with fragile structure \\[0.35em]

Diversity 
& \textbf{Internal and persistent}: alternatives coexist indefinitely, enabling implicit QD and cross-context evaluation
& External only: diversity exists across internally single-path individuals overwritten every iteration \\[0.35em]
Stability 
& High: strong paths preserved; repair ensures acyclicity and consistency 
& Low–medium: structural breakage common; no guaranteed repair \\[0.35em]

Partial Solution Reuse
& \textbf{Yes}: alternatives remain executable and recombine across paths 
& No: past variants are discarded and cannot be reused \\[0.35em]

Expressivity 
& Exponential combinatorial search space with operator-level statistics 
& Limited to sequential or small-population exploration \\

\bottomrule
\end{tabular}
\end{table}

\twocolumn

\section*{Appendix B: Minimal EvoLattice Example}
\label{app:minimal_example}

\begin{lstlisting}[basicstyle=\ttfamily\footnotesize]
spec_top1_vec:
- "lambda input:
    torch.stack([v[0] for v in input['spectral'].values()])
    if len(input['spectral']) > 1 else torch.zeros(2)
  # name: spec_top1_vec_0
  # mean=-0.0087 std=0.0312 max=0.0938 age=16"
- "lambda input:
    torch.nn.functional.normalize(
        torch.stack([v[0] for v in input['spectral'].values()]),
        p=2, dim=0)
    if len(input['spectral']) > 1 else torch.zeros(2)
  # name: spec_top1_vec_1
  # mean=-0.0076 std=0.0320 max=0.1103 age=7"
spectral_stability:
- "lambda spectral_cv_abs:
    1.0 / (1.0 + torch.abs(spectral_cv_abs))
  # name: spectral_stability_0
  # mean=-0.0065 std=0.0351 max=0.1542 age=4"
- "lambda spectral_cv_abs:
    1.0 / (1.0 + torch.abs(spectral_cv_abs).pow(0.5))
  # name: spectral_stability_1
  # mean=-0.0080 std=0.0295 max=0.1103 age=2"
zerolm_core:
- "lambda spec_topk_mean, spectral_stability, cov_sum:
    torch.tanh(spec_topk_mean)
    * (0.7 * spectral_stability + 0.3 * torch.sigmoid(cov_sum))
  # name: zerolm_core_0
  # mean=-0.0076 std=0.0307 max=0.0926 age=9"
- "lambda spec_topk_mean, spectral_stability, cov_sum:
    torch.tanh(spec_topk_mean)
    * (torch.sigmoid(spec_topk_mean) * spectral_stability
       + (1 - torch.sigmoid(spec_topk_mean))
         * torch.sigmoid(cov_sum))
  # name: zerolm_core_1
  # mean=-0.0087 std=0.0323 max=0.0685 age=4"
output:
- "lambda zerolm_core, spectral_entropy:
    zerolm_core + 0.1 * spectral_entropy
  # name: output_0
  # mean=-0.0051 std=0.0331 max=0.1542 age=9"
- "lambda zerolm_core:
    zerolm_core
  # name: output_1
  # mean=-0.0079 std=0.0310 max=0.0926 age=6"
\end{lstlisting}

\section*{Appendix C: The Best Proxy Discovered}
\label{app:best_proxy}

\begin{lstlisting}[language=Python, basicstyle=\ttfamily\small]
def evolattice_proxy(spectral, cov):
    # Spectral top-1 vector
    spec_vec = torch.stack([v[0] for v in spectral.values()])
    # Top-k robust aggregation
    topk = torch.topk(spec_vec, k=min(3, spec_vec.numel()), sorted=False).values
    spec_topk_mean = (0.6 * topk.mean() + 0.4 * topk.clamp(min=1e-6).log().mean().exp())
    # Spectral stability (CV-based)
    spec_mean = spec_vec.mean()
    spec_var = spec_vec.var(unbiased=False)
    spectral_cv = (spec_var + 1e-6).sqrt() / (spec_mean.abs() + 1e-3)
    reliability_gate = 0.7 / (1.0 + spectral_cv.abs())
    # Covariance aggregation
    reliability_gate += 0.3 * sum(cov.values()).log1p().sigmoid()
    # ZeroLM-repaired core
    core = spec_topk_mean.tanh() * reliability_gate
    # Entropy tie-breaker (low weight)
    p = torch.softmax(spec_vec, dim=0)
    return core - 0.1 * (p * (p + 1e-8).log()).sum()
\end{lstlisting}

\section*{Appendix D: Optimizer Update Discovery}
\label{app:optimizer_discovery}

To assess EvoLattice's generalization beyond proxy learning and architecture ranking, we evaluate it on a qualitatively different task: \emph{training-free optimizer update rule discovery}. The objective is to synthesize a single-step parameter update rule that improves validation loss when applied \emph{virtually} \citep{foret2021sam}, without performing full training or maintaining optimizer state. This task isolates the intrinsic quality of an update rule and differs fundamentally from the NAS proxy task in both structure and objective. A candidate update rule produces an update $\Delta w$ from gradients $g$ and a diagonal curvature proxy $h$. The update is applied virtually to model parameters, validation loss is measured, and parameters are immediately restored. All methods are evaluated under identical gradients, curvature estimates, and scoring weights. We report best-path performance only, consistent with the single-rule nature of the task, and to avoid redundancy with the main experimental analysis. We compare EvoLattice against strong handcrafted training-free optimizer update rules, including SGD, sign-based updates (SignSGD \& Lion), diagonal curvature-normalized updates, and manually designed sign--curvature hybrid baselines with linear or nonlinear gating. All baselines are instantiated from the same gradient and curvature primitives available, ensuring a fair comparison between manual composition and EvoLattice’s multi-path evolution.

\begin{figure*}[t]
    \centering
    \includegraphics[
        width=\textwidth,
        trim={0 0 0 0cm},
        clip
    ]{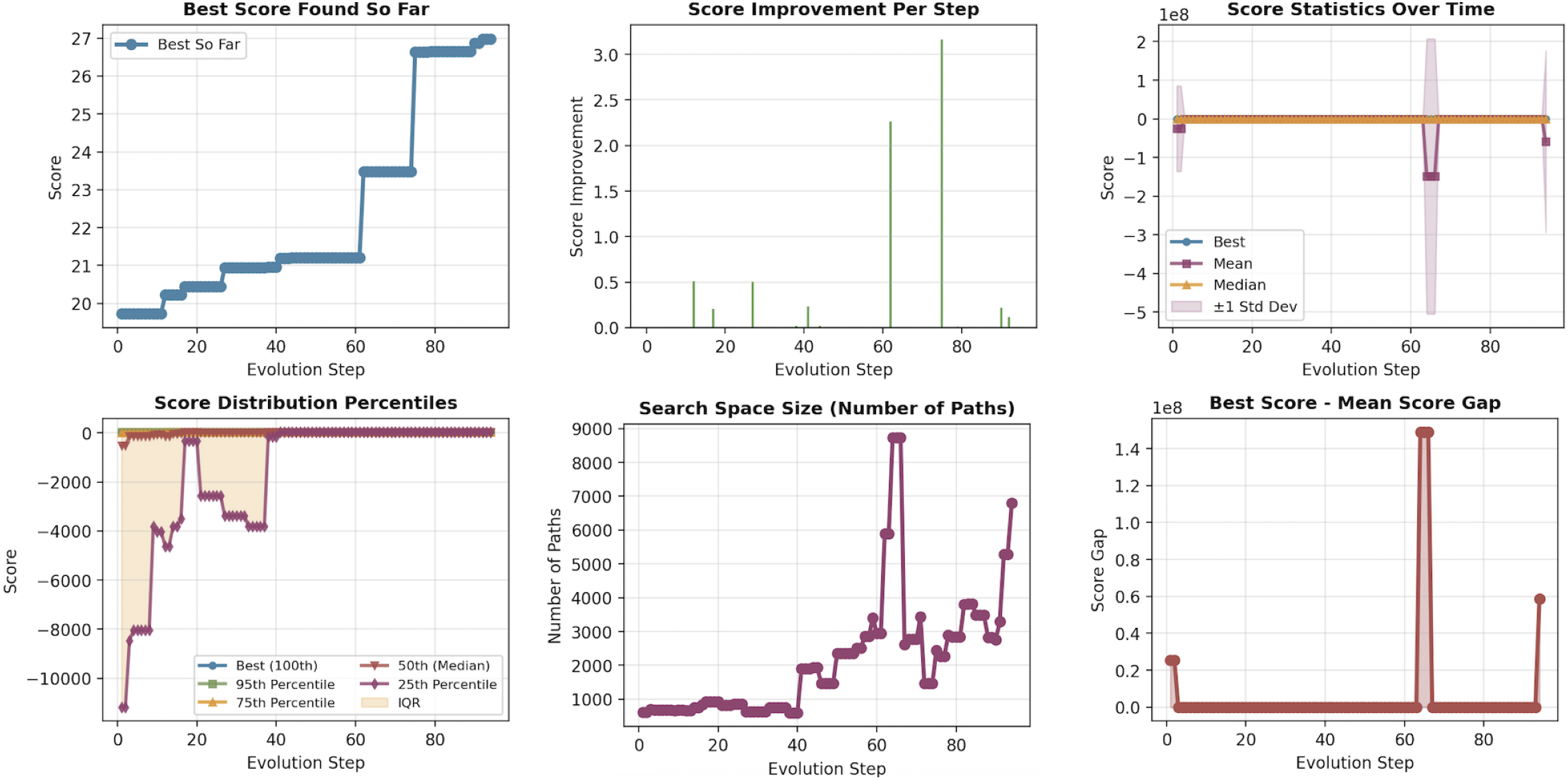}
    \caption{EvoLattice exhibits monotonic best-path improvement with sparse jumps,
    bounded variance despite rapid growth of the combinatorial search space,
    and a widening separation between elite and average paths in training-free optimizer discovery.}
    \label{fig:evolution_dynamics_appendix_2}
\end{figure*}

\begin{table}[t]
\centering
\footnotesize
\setlength{\tabcolsep}{3pt}
\renewcommand{\arraystretch}{0.95}
\caption{Best-path performance on the training-free optimizer discovery.
Scores measure validation loss improvement under a virtual parameter update,
combined with alignment and sharpness penalties.
All methods use the same gradient and curvature inputs.}
\label{tab:optimizer_results_app}
\begin{tabular}{@{}p{0.45\columnwidth} c p{0.32\columnwidth}@{}}
\toprule
\textbf{Method} & \textbf{Score ($\uparrow$)} & \textbf{Notes} \\
\midrule
SGD ($-g$) & 14.2 & First-order gradient descent \\
SignSGD \citep{bernstein2018signsgd} & 17.9 & Pure sign update \\
Lion \citep{chen2023lion} & 18.6 & SignSGD with scale modulation \\
Curvature-normalized ($-g/\sqrt{h}$) & 19.1 & Diagonal 2nd-order proxy \\
Sign + curvature\\ (linear blend) & 20.4 & Handcrafted hybrid baseline \\
Sign + curvature\\ (nonlinear gating) & 21.7 & Tuned hybrid baseline \\
\midrule
\textbf{EvoLattice (ours): nonlinear sign--curvature operator}
& \textbf{26.9} & \\
\bottomrule
\end{tabular}
\end{table}

EvoLattice achieves a substantially higher best-path score than all handcrafted training-free optimizer baselines constructed from the same primitives. As in the main NAS proxy experiment, the improvement reflects the quality of a \emph{single discovered update rule}, not an ensemble effect. This indicates that EvoLattice’s internal multi-alternative representation and alternative-level credit assignment remain effective in a non-proxy, optimizer-synthesis setting. Figure~\ref{fig:evolution_dynamics_appendix_2} visualizes EvoLattice’s internal evolution dynamics on this task. The best-path score improves monotonically via sparse, high-impact jumps, while the mean score across all executable paths remains near zero as exploratory alternatives persist. The gap between best and mean scores widens over time, variance remains bounded, and the number of executable paths grows rapidly. These dynamics mirror the quality--diversity behavior observed in the main experiment, and corroborate that EvoLattice’s internal population,
non-destructive mutation, and alternative-level credit assignment
are not specific to proxy discovery.
The same evolutionary dynamics enable the autonomous synthesis of
high-quality optimizer update rules, reinforcing its generality.

The highest-scoring path corresponds to a single emergent sign--curvature operator
rather than an explicit ensemble or linear blend.
Curvature information enters implicitly through nonlinear gating
rather than direct normalization.
A simplified, faithful form of the discovered update rule is:
\[
\begin{aligned}
\Delta w
&=
-\tanh\!\Bigl(\mathrm{sign}(g)\,\bigl(1 + \alpha\,\|w\|_1\bigr)\Bigr) \\
&\quad+\;\Bigl[
-\beta\,\sigma\!\Bigl(\tfrac{g}{\sqrt{h}+\epsilon}\Bigr)
+\gamma\,\mathrm{sign}(g)\,\bigl(1 + \delta\,
\mathrm{Var}(h)\bigr)
\Bigr]\,.
\end{aligned}
\]

where $\sigma(\cdot)$ denotes the sigmoid function and
$\alpha,\beta,\gamma,\delta > 0$ are constants selected through evolution.
This structure resembles modern sign-based optimizers,
but introduces implicit curvature-aware modulation via nonlinear gating.


\end{document}